%% file: template.tex
\documentclass{Interspeech}

\newcommand{\Sref}[1]{\S\ref{#1}}
\newcommand{\Fref}[1]{Figure~\ref{#1}}
\newcommand{\Tref}[1]{Table~\ref{#1}}
\usepackage{multirow}
\usepackage{tikz}
\usepackage{pgfplots, pgfplotstable}   
\usepackage{hyperref}

\interspeechcameraready

\title{CS-FLEURS: A Massively Multilingual and Code-Switched Speech Dataset}

\author[affiliation={1}]{Brian}{Yan}
\author[affiliation={2}]{Injy}{Hamed}
\author[affiliation={3}]{Shuichiro}{Shimizu}
\author[affiliation={4}]{Vasista}{Lodagala}
\author[affiliation={1}]{William}{Chen}
\author[affiliation={5}]{Olga}{Iakovenko}
\author[affiliation={6}]{Bashar}{Talafha}
\author[affiliation={7}]{Amir}{Hussein}
\author[affiliation={8}]{Alexander}{Polok}
\author[affiliation={1}]{Kalvin}{Chang}
\author[affiliation={8}]{Dominik}{Klement}
\author[affiliation={4}]{Sara}{Althubaiti}
\author[affiliation={9}]{Puyuan}{Peng}
\author[affiliation={7}]{Matthew}{Wiesner}
\author[affiliation={2}]{Thamar}{Solorio}
\author[affiliation={4}]{Ahmed}{Ali}
\author[affiliation={7}]{Sanjeev}{Khudanpur}
\author[affiliation={1}]{Shinji}{Watanabe}
\author{Chih-Chen}{Chen}
\author[affiliation={1}]{Zhen}{Wu}
\author[affiliation={9}]{Karim}{Benharrak}
\author[affiliation={9}]{Anuj}{Diwan}
\author[affiliation={1}]{Samuele}{Cornell}
\author[affiliation={9}]{Eunjung}{Yeo}
\author[affiliation={9}]{Kwanghee}{Choi}
\author{Carlos}{Carvalho}
\author[affiliation={1}]{Karen}{Rosero}

\affiliation{}{Carnegie Mellon University}{ $^2$Mohamed bin Zayed University of Artificial Intelligence, $^3$Kyoto University, $^4$Humain, $^5$University of Sheffield, $^6$ University of British Columbia, $^7$Johns Hopkins University, $^8$Brno University of Technology, $^9$University of Texas at Austin}


\usepackage{comment}
\usepackage{xcolor}

\makeatletter
\def\blfootnote{\xdef\@thefnmark{}\@footnotetext}
\makeatother

\begin{document}

\maketitle

\vspace*{1em}

\begin{abstract}    
    We present CS-FLEURS, a new dataset for developing and evaluating code-switched speech recognition and translation systems beyond high-resourced languages.
    CS-FLEURS consists of 4 test sets which cover in total 113 unique code-switched language pairs across 52 languages: 1) a 14 X-English language pair set with real voices reading synthetically generated code-switched sentences, 2) a 16 X-English language pair set with generative text-to-speech 3) a 60 \{Arabic, Mandarin, Hindi, Spanish\}-X language pair set with the generative text-to-speech, and 4) a 45 X-English lower-resourced language pair test set with concatenative text-to-speech.
    Besides the four test sets, CS-FLEURS also provides a training set with 128 hours of generative text-to-speech data across 16 X-English language pairs.
    Our hope is that CS-FLEURS helps to broaden the scope of future code-switched speech research.
    Dataset link: \url{https://huggingface.co/datasets/byan/cs-fleurs}.
\end{abstract}

\section{Introduction}

In the current era of massively multilingual speech processing, practitioners are developing and evaluating systems on hundreds of languages \cite{li2022asr2k,radford2023robust, zhang2023google, peng24b_interspeech, pratap2024scaling}.
This increasingly global impact is due in large part to dataset construction efforts, which have enabled pre-training on raw untranscribed speech \cite{kahn2020libri, wang2021voxpopuli} and supervised training on transcribed or pseudo-labeled speech \cite{panayotov2015librispeech, pratap2020mls, ardila2020common, li2023yodas}, as well as broad benchmarking on standardized evaluation sets across many languages \cite{shi2023ml, conneau2023fleurs}.
However, these data resources are collections of predominately \textit{monolingual} utterances - we currently lack broad \textit{code-switched} speech data, where utterances consist of more than one language.

Despite its prevalence in multilingual communities, conversational code-switched speech is particularly challenging to collect at scale. 
Unlike monolingual data, which can be sourced from a wide range of speakers, code-switched speech requires bilingual or multilingual speakers who naturally mix languages in conversation \cite{auer2013code}. 
Additionally, code-switching tends to occur sporadically and unpredictably, making it difficult to capture in large, curated datasets. 
Even when data is available, standardizing it across language pairs presents another challenge. 
These factors make it difficult to create a single, broadly representative dataset of code-switched speech; instead, practitioners have focused on constructing dedicated corpora for their particular communities of interest \cite{deuchar2014building, lyu2010seame, van-der-westhuizen-niesler-2018-first, chowdhury21_interspeech, diwan21_interspeech, hamed2022arzen}.

In this work, we take an alternative approach: rather than relying on the collection of natural conversational code-switched speech, we use a combination of read speech and synthetic speech over synthetically generated code-switched text.
Although our approach does not capture code-switched speech that occurs in conversation, we are able to use (1) \textbf{standard code-switching patterns}, (2) a \textbf{standard textual domain}, and (3) a \textbf{standard audio domain} across language pairs. 
By controlling for these naturally occurring variations when constructing the dataset, we can then ask: \textit{how do models perform on code-switched speech across different language pairs}?

We present CS-FLEURS: a massively multilingual and code-switched ASR and ST dataset consisting of 52 languages and 113 unique code-switched pairs across three subsets:
\begin{itemize}
    \item \textsc{cs-fleurs-read}: 14 X-English pairs, read speech 
    \item \textsc{cs-fleurs-xtts}: 76 pairs across 17 langs, generative TTS 
    \item \textsc{cs-fleurs-mms}: 45 X-English pairs, concatenative TTS 
\end{itemize}
To the best of our knowledge, this is the broadest single collection of code-switched speech data both in terms of languages covered and number of unique code-switched pairs.

\begin{figure}[t]
\vspace*{2em}
\centering
\includegraphics[width=\linewidth]{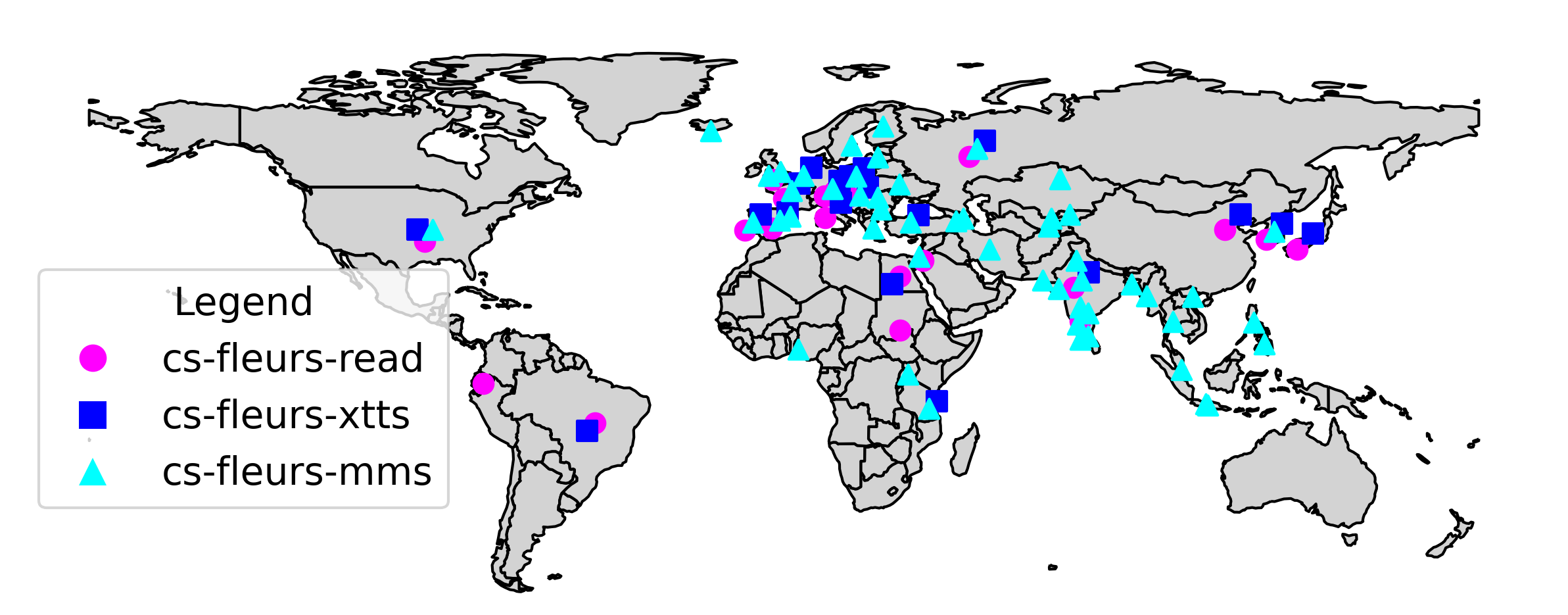}
\caption{Illustration of where the 52 languages covered by CS-FLEURS are spoken around the world.}
\label{fig:map}
\vspace{-2em}
\end{figure}

This dataset supports model benchmarking across 113 diverse pairs and model training across 16 X-English pairs.
\Fref{fig:map} illustrates the language coverage and \Tref{tab:data} summarizes the dataset with comparison to other code-switched corpora.
In addition to the dataset, this work also describes several empirical findings.
First, we show that Whisper \cite{radford2023robust} struggles with transcribing code-switched speech, especially for language pairs with distinct scripts.
However, we also show that Whisper is relatively robust at translating code-switched speech to English, suggesting that code-switched ST is an easier task than ASR.
Finally, we show that synthesizing code-switched speech data is an effective form of training data augmentation, improving model performance on both seen and unseen language pairs.

\section{Dataset}

\begin{table}[t]
  \centering
    \caption{CS-FLEURS vs other code-switched speech corpora.}
    \vspace{-3mm}
    \include{tables/data}
    \label{tab:data}
    \vspace{-2em}
\end{table}

CS-FLEURS builds upon a line of prior works.
The FLoRes-101 dataset \cite{goyal2022flores} consists of 3001 sentences from English Wikipedia translated in 101 languages by human translators.
Then, the FLEURS dataset \cite{conneau2023fleurs} took 2009 of those sentences from FLoReS-101 and recorded read speech in 102 languages with three speakers per language.
Now, CS-FLEURS takes the 2009 utterances from FLEURS to generate code-switched text and speech across 113 language pairs, keeping the same train/dev/test splits from FLEURS.

\begin{table}[t]
  \centering
    \caption{The 113 code-switched language pairs covered.}
    \vspace{-2mm}
    \include{tables/taxonomy}
    \label{tab:taxonomy}
    \vspace{-3em}
\end{table}

\begin{table}[t]
  \centering
    \caption{Summary statistics for CS-FLEURS.}
    \vspace{-3mm}
    \include{tables/stats}
    \label{tab:stats}
    \vspace{-3em}
\end{table}

We follow the Matrix Language-Frame model \cite{myers1997duelling}; for each code-switched pair, one language, referred to as Matrix, provides grammatical structure while a second language, referred to as Embedded, provides words or morphological units that are inserted within the sentence. We refer to language pairs as Matrix-Embedded; for instance, Mandarin-English refers to a Mandarin matrix sentence with English words embedded.
\Tref{tab:taxonomy} shows a breakdown of the code-switched language pairs covered by CS-FLEURS across the \textsc{cs-fleurs-read}, \textsc{cs-fleurs-xtts}, and \textsc{cs-fleurs-mms} subsets.
\Tref{tab:stats} provides additional summary statistics.

\subsection{Collecting Read Speech Across 14 Language Pairs}

\textsc{cs-fleurs-read} consists of read speech from bilingual speakers across 14 X-English language pairs (1-4 speakers per language pair, 2:1 M:F ratio overall).
\textsc{cs-fleurs-read} is generated using the test set of FLEURS and is intended for model benchmarking. 
As such, this set has been human validated.
In terms of language coverage, the limiting factor for constructing this set was the availability of bilingual readers.
Finally, in this set we regard the 14 non-English languages as Matrix and English as Embedded. 

\subsubsection{Generating accurate and fluent code-switched text}
\label{sec:llm}

We use a large language model (LLM) backbone for generating code-switched text for \textsc{cs-fleurs-read}.
Specifically, by utilizing X-English parallel monolingual sentences from FLEURS test set, we prompt GPT-4o \cite{hurst2024gpt} to generate the equivalent code-switched sentences in three different manners:
\begin{itemize}
    \item \textit{GPT-Base}: feeding only the paired monolingual sentences
    \item \textit{GPT-EC}: 
    feeding paired monolingual sentences and a set of valid English words to be embedded, determined by the Equivalence Constraint theory \cite{poplack1980EC} so as not to violate the syntactic rules of both languages, as implemented in \cite{kuwanto2024linguistics}
    \item \textit{GPT-Pred}: feeding paired monolingual sentences and a set of valid English words to be embedded, determined by a predictive model \cite{hamed2023investigating}, which given an English sentence, predicts the plausible English words to be embedded in a corresponding code-switched sentence
\end{itemize}
\Tref{tab:prompts} shows an example of a full prompt for a given sentence pair.
\textit{GPT-EC} and \textit{GPT-Pred} prompts include English keywords for embedding into the matrix language.
Further, to encourage morphological code-switching, where more than one language occurs \textit{within the same word}, we prompt GPT-4o with an in-context example, as illustrated in the table. 10 out of 14 languages were prompted as such.\footnote{10 w/ morphological example: ara, ces, deu, ita, kor, por, rus, slk, spa, tel, tam; 4 w/o: cmn, fra, hin, jpn}

The resultant generations are not guaranteed to be (1) code-switched, (2) accurate, preserving the meaning of the original monolingual sentences, or (3) fluent, plausibly produced by a human.
We ask our bilingual readers to reject sentences which are not code-switched or not accurate.
Sentences are not rejected on fluency, but we collect ratings from 0-2, with 0 indicating unnaturalness, 1 indicating somewhat natural, and 2 indicating perfect naturalness.

 \begin{table}[t]
   \centering
     \caption{The prompt consists of three main parts: (1) task description and the paired monolingual sentences to be synthesized (outlined in black); (2) the set of valid English words in the case of \textit{GPT-EC} and \textit{GPT-Pred} (outlined in \textcolor{blue}{blue}); and (3) morphological code-switching example in the case of languages
     where this type of switching is common (outlined in \textcolor{violet}{purple}).}
     \vspace{-2mm}
     \include{tables/prompt_general}
     \label{tab:prompts}
 \end{table}

\begin{table}[t]
  \centering
    \caption{Human validation on LLM-generated code-switched text across 14 X-English pairs. CMI=Code-Mixing Index.}
    \vspace{-2mm}
    \include{tables/validation}
    \label{tab:validation}
    \vspace{-3em}
\end{table}

As shown in \Tref{tab:validation}, the reject rate was fairly consistent across the different prompts.
In terms of fluency, \textit{GPT-Pred} scored highest while \textit{GPT-EC} was lowest; this result correlates with the frequency of switching, as measured by the Code-Mixing Index (CMI) \cite{gamback2014comparing}.
We also noted that \textit{GPT-EC} embeds function words a high rate, which bilingual readers found to be rather unnatural. 
\textit{GPT-Pred} on the other hand primarily embeds content words, which resulted in over over 80\% of generations to be rated somewhat or perfectly natural.

\subsubsection{Reading and recording code-switched speech}

To facilitate the collection of read speech, we developed an open-source toolkit.\footnote{\url{https://github.com/brianyan918/sentence-recorder/tree/codeswitching}}
We distributed generated code-switched text to 21 bilingual readers.
For each sentence, we randomly selected one of the three aforementioned prompts, standardizing the selection across all language pairs.
The bilingual readers were provided with the monolingual references along with the code-switched sentences.
They were instructed to 1) verify that the sentence is indeed code-switched, 2) the sentence has the same general meaning as the monolingual reference, 3) read and record the sentence, and 4) rate the fluency of the sentence.
Recordings were then validated by a human listener.

\subsection{Synthesizing Speech Across 111 Language Pairs}

Next, we describe two synthetic speech sets designed to complement \textsc{cs-fleurs-read} by covering additional language pairs and providing code-switched training data.

\textsc{cs-fleurs-xtts} consists of synthetic speech from the XTTS-v2 model \cite{casanova2024xtts}, which supports 17 languages, across 16 X-English language pairs as well as 60 non-English language pairs (15 Arabic-X, 15 Hindi-X, 15 Mandarin-X, 15 Spanish-X).
For the 16 X-English language pairs, we generate train, dev, and test sets, regarding the 16 non-English languages as matrix and English as embedded --- 12 of these language pairs overlap with \textsc{cs-fleurs-read}.
The 60 non-English language pairs are test-only and we regard Arabic, Hindi, Mandarin, or Spanish as the matrix and the other language as embedded --- none of these language pairs overlap with the other two subsets.

\textsc{cs-fleurs-mms} consists of synthetic speech from MMS TTS models \cite{pratap2024scaling} across 45 X-English language pairs, many of which are lower-resourced --- only 10 of these language pairs overlap with the other two subsets.
This set is test-only and we regard the non-English languages as matrix.

Both sets are not human validated due to lack of bilingual speakers. Instead, we use language-universal forced alignment to filter low quality generations.

\subsubsection{Cheaply and flexibly generating code-switched text}

While the LLM backbone described in \Sref{sec:llm} can generate natural and morphologically rich code-switching, there are also a number of drawbacks that limit scalability aside from the high computational (or API) cost.
The LLM-based methods required human validation to rejected around 10\% of the sentences across 14 X-English language pairs; it is reasonable to expect a higher reject rate for lower resourced languages.

Further, we found that the LLM-based methods produced a wide range of code-switching frequency, as measured by CMI \cite{gamback2014comparing}, across language pairs.
This variance is shown in the first line of \Tref{tab:cmi}: the standard deviation of the CMI across 12 language pairs in the human validated \textsc{cs-fleurs-read} is 10.8.
We therefore opt for a more simple and rigid method for generating the code-switched text in \textsc{cs-fleurs-xtts} and \textsc{cs-fleurs-mms}, referred to as \textit{align-then-swap}.

The align-then-swap procedure is simple.
We first obtain word-level alignment using AwesomeAlign \cite{dou2021word}, a 104 language mBERT model \cite{devlin-etal-2019-bert} fine-tuned towards word alignment objectives using parallel text across five language pairs.
Notably, AwesomeAlign generalizes to other languages that were part of the mBERT pre-training but unseen during the word alignment fine-tuning stage \cite{dou2021word}.
Next we randomly select 30\% nouns, verbs, abverbs, and adjectives, tagged using Stanza \cite{qi2020stanza}, to be swapped with the aligned embedded language words.

Additionally, we take care of one-to-many word alignments (matrix-to-embedded) by inserting words in the order that they originally appeared in the monolingual embedded language sentence.
We also take care of character-based languages (Mandarin and Japanese) by first re-segmenting to words using Stanza \cite{qi2020stanza}; Korean text in FLEURS was already word segmented, so this extra step was not necessary.

\begin{table}[t]
  \centering
    \caption{Comparison of the code-switching frequency, as measured by Code-Mixing Index (CMI), of text generated by LLM prompts and align-then-swap (Swap) across 12 X-English language pairs. M=Mean; SD=Standard Deviation.}
    \vspace{-2mm}
    \include{tables/cmi}
    \label{tab:cmi}
    \vspace{-3em}
\end{table}

As shown in \Tref{tab:cmi}, the resultant CMI from align-then-swap (Swap) is more consistent across language pairs than its LLM-based counterpart: the standard deviation is only 4.0, which is 60\% lower.
This simpler, yet more consistent, method of generating code-switched text is therefore preferable for the long tail of lower resourced languages and rare language pairs.

\subsubsection{Generative code-switched synthesis with XTTS}

XTTS-v2 \cite{casanova2024xtts} is a multilingually trained TTS model based on a GPT-2 encoder that predicts VQ-VAE units which then latently condition a HiFi-GAN decoder.
The model was trained on 17 languages, which are distinguished by appending a language ID token (e.g. [en]) at the start of textual inputs.
XTTS-v2 also incorporates a voice conversion component by conditioning both the GPT-2 encoder and VQ-VAE decoder on speaker information; in practice only single utterance from the target speaker is necessary to enable voice conversion.

To synthesize code-switched speech using XTTS-v2, we feed generated code-switched text to the model with the matrix language ID token.
Mandarin, Japanese, and Korean text is romanized, in the same manner as the original XTTS-v2 training.
We use a matrix language speaker from the FLEURS dataset for the voice conversion component.
XTTS-v2 produces 24khz speech, which we then down-sample to 16khz.

\subsubsection{Concatenative code-switched synthesis with MMS-TTS}

To cover additional language pairs, we use a concatenative approach that generates monolingual TTS in segments using MMS-TTS \cite{pratap2024scaling}.
We concatenate the monolingual segments, inserting 100ms of silence in between.
Since MMS-TTS is a single-speaker VITS style model, the resulting concatenated code-switched speech contains unnatural artifacts and speaker changes; however the content is correct.
Unlike the XTTS-v2 outputs, MMS-TTS outputs do not contain any accented speech.

\subsubsection{Universal forced alignment filtering}

\begin{table}[t]
  \centering
    \caption{Comparison of the filtered vs accepted synthetic speech. Filtering was done by forced alignment score (FAS). 
    }
    \vspace{-2mm}
    \include{tables/filtering}
    \label{tab:filtering}
    \vspace{-3em}
\end{table}

To perform quality control, we use a universal forced alignment model, MMS-ZS \cite{zhao2024scaling}, which was trained to align speech in thousands of languages to romanized text - this model provides a language-universal measure of intelligibility.
Utterances with the 5\% lowest length-normalized forced alignment score (FAS) are filtered within each language pair.

In \Tref{tab:filtering}, we report the average FAS of the filtered vs accepted portions of \textsc{xtts-test1} and \textsc{mms-test}, along with Whisper-Large-v3 \cite{radford2023robust} character error rate (CER), the UTMOS (1-5) \cite{saeki2022utmos} automatic naturalness metric, and the number of speakers changes identified by the Pyannote-3.1 \cite{Bredin23} diarization model (SCD).
For \textsc{xtts-test1}, filtering has a large effect on Whisper CER and a moderate effect on UTMOS, indicating that the generative synthesis method is prone to producing unintelligible and unnatural speech.
On the other hand for \textsc{mms-test}, filtering has less effect on CER and UTMOS, indicating that concatenative synthesis is less error prone; however, SCD shows that this method results in frequent speaker changes.

\section{Experiments}

In this section, we describe several empirical findings based on CS-FLEURS.
For ASR, we measure case insensitive and unpunctuated character error rate (CER).
For ST (to English), we measure case insensitive and unpunctuated BLEU \cite{post-2018-call}.
\vspace{-1mm}
\subsection{Benchmarking}
\vspace{-1mm}

\begin{figure}[t]
\centering
\include{figures/fl_vs_cs-fl}
\vspace{-8mm}
\caption{Whisper-Large-v3 ASR (CER$\downarrow$) and ST to English (BLEU$\uparrow$) on CS-FLEURS test sets (\textcolor{blue}{blue}). Performance on two monolingual control sets are shown for comparison (gray).}
\vspace{-4mm}
\label{fig:fl_vs_cs-fl}
\end{figure}

\textit{How does Whisper perform on code-switched speech as compared to monolingual speech?}
To answer this, we compare ASR and ST (to English) performance on each CS-FLEURS test set with two monolingual control sets consisting of monolingual speech from all matrix languages in each respective CS-FLEURS set: 1) the original FLEURS and 2) a TTS version of FLEURS, which is monolingual.
The latter, referred to as TTS FLEURS, allows us to compare results on TTS-based code-switched speech with the matching TTS-based monolingual speech, eliminating the possibility that Whisper simply is not robust to XTTS or MMS-TTS speech.

As shown in \Fref{fig:fl_vs_cs-fl}, ASR CER is over 2x higher on CS-FLEURS than the original monolingual FLEURS - this degradation is largest on the non-English language pair set (\textsc{xtts-test2}).
However, the discrepancy between ST performance on monolingual and code-switched speech is low for XTTS and MMS sets.
Further, Whisper actually performs better on \textsc{read-test} than the original FLEURS - this can be attributed to the high rate of X-English code-switching (see \Tref{tab:cmi}).
These results indicate that \textbf{directly translating code-switched speech is less erroneous than relying on transcription}, although we leave benchmarking ST for non-English targets to future work.

\begin{table}[t]
  \centering
    \caption{Comparison of Whisper-Large-v3 ASR (CER$\downarrow$) performance on same-script vs distinct-script language pairs.}
    \vspace{-3mm}
    \include{tables/script}
    \label{tab:script}
    \vspace{-1em}
\end{table}

\textit{How does Whisper perform on code-switched speech across different language pairs?}
As shown in \Tref{tab:script}, performance on distinct-script language pairs is significantly worse than same-script language pairs - on average CER is 3x higher.
These results suggest that \textbf{code-switched ASR performance is limited by the ability to interchangeably produce two distinct scripts} within a single utterance-level decoding.

\vspace{-1mm}
\subsection{Training}
\vspace{-1mm}
\begin{table}[t]
  \centering
    \caption{Results on the effect of training data augmentation.}
    \vspace{-3mm}
    \include{tables/training}
    \label{tab:training}
    \vspace{-6mm}
\end{table}

Finally, we investigate the question: \textit{how useful is synthetic data for training models?}
To answer this, we train two models using the ESPnet toolkit \cite{watanabe18_interspeech}: one using the original FLEURS training data and a second using the original FLEURS plus CS-FLEURS (\textsc{xtts-train}).
Both models are trained for the same number of iterations following the self-conditioned XLSR-based recipe described in \cite{chen2023improving}.
We then evaluated on \textsc{read-test}, which consists of 14 language pairs; 12 of these are seen in \textsc{xtts-train} and 2 are unseen.
We also report the monolingual FLEURS result for the matrix languages of these 12 seen and 2 unseen language pairs.
As shown in \Tref{tab:training}, \textbf{training on synthetic code-switched data improved performance on both seen and unseen language pairs.}

\vspace{-1mm}
\section{Conclusion}
\vspace{-1mm}
CS-FLEURS is a hybrid read/synthetic speech dataset for developing and evaluating code-switched systems across 52 languages and 113 unique code-switched language pairs.
By using this dataset, which controls for in code-switching patterns, text domain, and audio domain, we find that transcribing speech across distinct-script language pairs remains difficult.

\pagebreak

\bibliographystyle{IEEEtran}
\bibliography{mybib}

\end{document}

%% file: tables/data.tex
\resizebox {\linewidth} {!} {
\setlength{\tabcolsep}{1.0pt}
\begin{tabular}{lccccccc}
\toprule
Dataset & Langs & CS-Pairs & Domain & Speech Type & Hours \\
\midrule
ArzEn \cite{hamed2022arzen} & 2 & 1 & Conversation & Spontaneous & 12 \\
Bangor Miami \cite{deuchar2014building} & 2 & 1 & Conversation & Spontaneous & 35 \\
SEAME \cite{lyu2010seame} & 2 & 1 & Conversation & Spontaneous & 192 \\
ESCWA \cite{chowdhury21_interspeech} & 3 & 2 & Conversation & Spontaneous & 3 \\
MUCS \cite{diwan21_interspeech} & 3 & 2 & Lecture & Structured & 160 \\
Soapies \cite{van-der-westhuizen-niesler-2018-first} & 5 & 4 & TV & Scripted & 14 \\
\midrule
\textbf{CS-FLEURS} & \textbf{52} & \textbf{113} & \textbf{Wikipedia} & \textbf{Read/Synthetic} & \textbf{294} \\

\bottomrule
\end{tabular}
}

%% file: tables/taxonomy.tex
\resizebox {\linewidth} {!} {
\setlength{\tabcolsep}{1.0pt}
\begin{tabular}{c|>{\centering\arraybackslash}m{3cm}|>{\centering\arraybackslash}m{3cm}|c|c|c}
\toprule
CS-Pairs & Matrix & Embedded & Read & XTTS & MMS \\
\midrule
7 & {ara deu fra hin por rus spa} & {eng} & \checkmark & \checkmark & \checkmark \\
\midrule
5 & {ces cmn ita jpn kor} & {eng} & \checkmark & \checkmark & \\
\midrule
2 & {slk tel} & {eng} & \checkmark & & \\
 \midrule
3 & {nld pol tur} & {eng} & & \checkmark & \checkmark \\
\midrule
1 & {hun} & {eng} & & \checkmark & \\
\midrule
60 & {ara cmn hin spa} & {ara fra deu ita por pol tur rus nld ces spa cmn jpn hun kor hin} & & \checkmark & \\
\midrule
35 & ben bul cat ceb cym ell fin guj heb hun ind isl jav kan kaz kir lav lug mal mar mya pan ron swe swh tam tel tgk tgl tha ukr urd uzb yor zlm & eng & & & \checkmark \\
\bottomrule
\end{tabular}
}

%% file: tables/stats.tex
\resizebox {\linewidth} {!} {
\setlength{\tabcolsep}{3pt}
\begin{tabular}{l|c|cccc|c}
\toprule
& \underline{\textsc{Read}} & \multicolumn{4}{c|}{\underline{\textsc{XTTS}}} & \underline{\textsc{MMS}} \\
Statistic & Test & Train & Dev & Test1 & Test2 & Test \\
\midrule
Duration (hours) & 17 & 128 & 15 & 36 & 42 & 56 \\
Tokens (words) & 128k & 889k & 105k & 257k & 300k & 315k \\
Matrix Langs & 14 & 16 & 16 & 16 & 4 & 45 \\
Embedded Langs & 1 & 1 & 1 & 1 & 15 & 1 \\
Total CS Pairs & 14 & 16 & 16 & 16 & 60 & 45 \\
Same-Script Pairs & 7 & 10 & 10 & 10 & 9 & 22 \\
Distinct-Script Pairs & 7 & 6 & 6 & 6 & 51 & 23 \\
\bottomrule
\end{tabular}
}

%% file: tables/prompt_general.tex
\footnotesize
\resizebox {\linewidth} {!} {
\setlength{\tabcolsep}{2.5pt}
\begin{tabular}{p{\linewidth}}
\toprule
You are a bilingual Spanish-English speaker, you will help translate Spanish and English sentences into a code-mixed sentence. The code-mixed sentence should contain words from both languages, each written in the correct language script. We will provide you with both Spanish and English sentences \textcolor{blue}{in addition to some English keywords that need to be present in the produced code-mixed sentence}. \\
\textcolor{violet}{You need to produce morphological code-switching when appropriate to achieve better fluency. The following is an example: }\\
\textcolor{violet}{Spanish: Mi amigo se enojó cuando le revelaron el final de la película.}\\
\textcolor{violet}{English: My friend freaked out when someone spoiled the end of the movie.} \textcolor{violet}{English keywords: freak out, spoiled}\\
\textcolor{violet}{A fluent code-mixed sentence would be: Mi amigo se frickeo cuando le spoilearon el final de la película.}\\
Please produce the code-mixed sentence for the following:\\
Spanish sentence: \{Spanish\_sentence\}\\
English sentence: \{English\_sentence\}\\
\textcolor{blue}{English keywords: \{English\_keywords\}}\\
Only provide the code-mixed sentence without explanation or extra text.\\
\bottomrule
\end{tabular}
}

%% file: tables/validation.tex
\resizebox {.9\linewidth} {!} {
\setlength{\tabcolsep}{3pt}
\begin{tabular}{l|c|cccc|c}
\toprule
& & \multicolumn{4}{c|}{\underline{Fluency}} &  \\
Prompt & Reject(\%) & 0(\%) & 1(\%) & 2(\%) & Avg & CMI  \\
\midrule
\textit{GPT-Base} & \textbf{8.54} & 31.66 & 25.44 & 34.36 & 1.03 & 25.50 \\
\textit{GPT-EC} & 12.15 & 44.71 & 21.22 & 21.92 & 0.74 & 30.98 \\
\textit{GPT-Pred} & 11.76 & 19.26 & 25.25 & 43.73 & \textbf{1.28} & 21.05 \\
\bottomrule
\end{tabular}
}

%% file: tables/cmi.tex
\resizebox {\linewidth} {!} {
\setlength{\tabcolsep}{1.2pt}
\begin{tabular}{l|cccccccccccc|cc}
\toprule
Type & ara & ces & cmn & deu & fra & hin & ita & jpn & kor & por & rus & spa & M & SD \\
\midrule
LLM & 12.3 & 30.4 & 24.1 & 27.8 & 36.0 & 18.5 & 37.0 & 18.5 & 7.0 & 37.2 & 17.7 & 39.8 & 25.5 & 10.8 \\
Swap & 20.4 & 20.1 & 14.11 & 16.8 & 15.0 & 13.6 & 16.8 & 11.2 & 25.9 & 17.0 & 20.6 & 15.5 & 17.2 & \textbf{4.0}\\
\bottomrule
\end{tabular}
}

%% file: tables/filtering.tex
\resizebox {\linewidth} {!} {
\setlength{\tabcolsep}{2.5pt}
\begin{tabular}{l|cccc|cccc}
\toprule
& \multicolumn{4}{c|}{\underline{\textsc{XTTS-Test1}}} & \multicolumn{4}{c}{\underline{\textsc{MMS-Test}}} \\
Subset & FAS$\uparrow$ & CER$\downarrow$ & UTMOS$\uparrow$ & SCD$\downarrow$ & FAS$\uparrow$ & CER$\downarrow$ & UTMOS$\uparrow$ & SCD$\downarrow$ \\
\midrule
Filtered & -0.68 & 32.95 & 2.46 & 1.12 & -0.96 & 43.10 & 3.02 & 1.90 \\
Accepted & -0.26 & 18.46 & 2.64 & 1.07 & -0.47 & 40.07 & 3.04 & 1.87 \\
\bottomrule
\end{tabular}
}

%% file: figures/fl_vs_cs-fl.tex
\resizebox {\linewidth} {!} {
\begin{tikzpicture}

  \node[rotate=90, anchor=center] at (-0.03\textwidth, 0.1\textwidth) {\Huge ASR};
  \node[rotate=90, anchor=center] at (-0.03\textwidth, -0.11\textwidth) {\Huge ST};

  \begin{scope}[xshift=0cm, yshift=0cm]
    \begin{axis}[
      ybar,
      width=0.38\textwidth,
      height=0.3\textwidth,
      xmin=0.5,
      xmax=1.5,
      ymin=0,
      ymax=60,
      ytick=\empty,
      xtick={0},
      bar width=34,
      title=\Large \textsc{READ-TEST},
      nodes near coords,
      every node near coord/.append style={font=\Large},
    ]
    \addplot[draw=black,fill=lightgray] coordinates { (1, 8.44) };
    \addplot[draw=black,fill=blue] coordinates { (1, 19.83) };
    \end{axis}
  \end{scope}
  
  \begin{scope}[xshift=0.3\textwidth, yshift=0cm]
    \begin{axis}[
      ybar,
      width=0.38\textwidth,
      height=0.3\textwidth,
      xmin=0.5,
      xmax=1.5,
      ymin=0,
      ymax=60,
      ytick=\empty,
      xtick={0},
      bar width=34,
      title=\Large \textsc{XTTS-TEST1},
      nodes near coords,
      every node near coord/.append style={font=\Large},
    ]
    \addplot[draw=black,fill=lightgray] coordinates { (1, 5.71) };
    \addplot[draw=black,fill=darkgray] coordinates { (1, 7.56) };
    \addplot[draw=black,fill=blue] coordinates { (1, 18.46) };
    \end{axis}
  \end{scope}

  \begin{scope}[xshift=0.6\textwidth, yshift=0cm]
    \begin{axis}[
      ybar,
      width=0.38\textwidth,
      height=0.3\textwidth,
      xmin=0.5,
      xmax=1.5,
      ymin=0,
      ymax=60,
      ytick=\empty,
      xtick={0},
      bar width=34,
      title=\Large \textsc{XTTS-TEST2},
      nodes near coords,
      every node near coord/.append style={font=\Large},
    ]
    \pgfkeys{/pgf/number format/.cd,fixed,fixed zerofill,precision=2}
    \addplot[draw=black,fill=lightgray] coordinates { (1, 12.50) };
    \addplot[draw=black,fill=darkgray] coordinates { (1, 14.26) };
    \addplot[draw=black,fill=blue] coordinates { (1, 36.06) };
    \end{axis}
  \end{scope}

  \begin{scope}[xshift=0.9\textwidth, yshift=0cm]
    \begin{axis}[
      ybar,
      width=0.38\textwidth,
      height=0.3\textwidth,
      xmin=0.5,
      xmax=1.5,
      ymin=0,
      ymax=60,
      ytick=\empty,
      xtick={0},
      bar width=34,
      title=\Large \textsc{MMS-TEST},
      nodes near coords,
      every node near coord/.append style={font=\Large},
    ]
    \pgfkeys{/pgf/number format/.cd,fixed,fixed zerofill,precision=2}
    \addplot[draw=black,fill=lightgray] coordinates { (1, 22.03) };
    \addplot[draw=black,fill=darkgray] coordinates { (1, 22.33) };
    \addplot[draw=black,fill=blue] coordinates { (1, 40.07) };
    \end{axis}
  \end{scope}

  \begin{scope}[xshift=0cm, yshift=-0.22\textwidth]
    \begin{axis}[
      ybar,
      width=0.38\textwidth,
      height=0.3\textwidth,
      xmin=0.5,
      xmax=1.5,
      ymin=0,
      ymax=60,
      ytick=\empty,
      xtick={0},
      bar width=34,
      nodes near coords,
      every node near coord/.append style={font=\Large},
    ]
    \addplot[draw=black,fill=lightgray] coordinates { (1, 30.23) };
    \addplot[draw=black,fill=blue] coordinates { (1, 47.27) };
    \end{axis}
  \end{scope}
  
  \begin{scope}[xshift=0.3\textwidth, yshift=-0.22\textwidth]
    \begin{axis}[
      ybar,
      width=0.38\textwidth,
      height=0.3\textwidth,
      xmin=0.5,
      xmax=1.5,
      ymin=0,
      ymax=60,
      ytick=\empty,
      xtick={0},
      bar width=34,
      nodes near coords,
      every node near coord/.append style={font=\Large},
      legend columns=3,
      legend style={
        at={(1.1,-0.1)}, 
        anchor=north, 
        draw=none, 
        fill=none, 
        column sep=0.4cm, 
        legend cell align=left 
       },
       legend image code/.code={
        \draw[#1] (0cm,-0.2cm) rectangle (0.4cm,0.4cm);
       },
    ]
    \addlegendentry{\Large \textsc{ORIGINAL FLEURS}}
    \addlegendentry{\Large \textsc{TTS FLEURS}}
    \addlegendentry{\Large \textsc{CS-FLEURS}}
    \addplot[draw=black,fill=lightgray] coordinates { (1, 28.31) };
    \addplot[draw=black,fill=darkgray] coordinates { (1, 27.04) };
    \addplot[draw=black,fill=blue] coordinates { (1, 23.81) };
    \end{axis}
  \end{scope}

  \begin{scope}[xshift=0.6\textwidth, yshift=-0.22\textwidth]
    \begin{axis}[
      ybar,
      width=0.38\textwidth,
      height=0.3\textwidth,
      xmin=0.5,
      xmax=1.5,
      ymin=0,
      ymax=60,
      ytick=\empty,
      xtick={0},
      bar width=34,
      nodes near coords,
      every node near coord/.append style={font=\Large},
    ]
    \pgfkeys{/pgf/number format/.cd,fixed,fixed zerofill,precision=2}
    \addplot[draw=black,fill=lightgray] coordinates { (1, 24.30) };
    \addplot[draw=black,fill=darkgray] coordinates { (1, 26.35) };
    \addplot[draw=black,fill=blue] coordinates { (1, 20.66) };
    \end{axis}
  \end{scope}

  \begin{scope}[xshift=0.9\textwidth, yshift=-0.22\textwidth]
    \begin{axis}[
      ybar,
      width=0.38\textwidth,
      height=0.3\textwidth,
      xmin=0.5,
      xmax=1.5,
      ymin=0,
      ymax=60,
      ytick=\empty,
      xtick={0},
      bar width=34,
      nodes near coords,
      every node near coord/.append style={font=\Large},
    ]
    \pgfkeys{/pgf/number format/.cd,fixed,fixed zerofill,precision=2}
    \addplot[draw=black,fill=lightgray] coordinates { (1, 23.89) };
    \addplot[draw=black,fill=darkgray] coordinates { (1, 21.80) };
    \addplot[draw=black,fill=blue] coordinates { (1, 16.58) };
    \end{axis}
  \end{scope}

\end{tikzpicture}
}

%% file: tables/script.tex
\resizebox {.8\linewidth} {!} {
\begin{tabular}{l|c|cc|c}
\toprule
& \underline{\textsc{Read}} & \multicolumn{2}{c|}{\underline{\textsc{XTTS}}} & \underline{\textsc{MMS}} \\
CS Pair Type & Test & Test1 & Test2 & Test \\ 
\midrule
Same Script & 7.32 & 8.49 & 9.92 & 28.26\\
Distinct Script & 32.33 & 37.17 & 40.67 & 51.62 \\
\bottomrule
\end{tabular}
}

%% file: tables/training.tex
\resizebox {.9\linewidth} {!} {
\begin{tabular}{c|cc|cc}
\toprule
Data & \multicolumn{2}{c|}{\underline{\textsc{fleurs}}} & \multicolumn{2}{c}{\underline{\textsc{cs-fleurs-read}}} \\
Augment & 12 Seen & 2 Unseen & 12 Seen & 2 Unseen \\
\midrule
 & 14.38 & 8.93 & 31.77 & 29.62 \\
\checkmark & \textbf{12.67} & \textbf{8.51} & \textbf{26.24} & \textbf{27.77} \\
\bottomrule
\end{tabular}
}